\pdfoutput=1

\documentclass[11pt]{article}

\usepackage[final]{acl}

\usepackage{times}
\usepackage{latexsym}

\usepackage[T1]{fontenc}

\usepackage[utf8]{inputenc}

\usepackage{microtype}

\usepackage{inconsolata}

\usepackage{graphicx}
\usepackage{hyperref}
\usepackage{amsmath}
\usepackage{amsfonts}
\usepackage[curve]{xypic}

\title{Cluster automata}

\author{Andr\'as Kornai\\
  Budapest University of Technology and Economics\\
  H-1111 Budapest Egry J u 1\\
  \texttt{kornai@math.bme.hu}}

\begin{document}
\maketitle
\begin{abstract}
We introduce a new class of {\it clustered} Moore automata (CMA), investigate
their temporal behavior, and describe some applications.
\end{abstract}

\noindent
Clustered Moore automata (CMA) are subsequential Moore transducers whose
states can contain smaller CMA that operate on a faster timescale, subject to
an Artinian limitation. In Section~\ref{sec:intro} we begin with some simple
examples, both to motivate the subsequent definitions and in order to fix
terminology and notation. In Section~\ref{sec:scale} we discuss the temporal
structures carried by CMAs in the finite domain that is of chief interest in
linguistic applications. Some of these applications, in particular island
parsing and disambiguation are described in Section~\ref{sec:ling}.

\section{Introduction}\label{sec:intro}

We will use Moore (as opposed to Mealy) automata, but keep our terminology
neutral across deterministic and nondeterministic, full and partial flavors,
and whether we use an {\it output function} $G:Q \rightarrow O^*$ where $Q$ is
the {\it state space} and $O$ is a finite {\it output alphabet} that may be
(but need not be) disjoint from the {\it input alphabet}
$\Sigma$. Terminological sticklers can call these `subsequential transducers'
rather than `automata' but many of our examples come from domains where
automata are naturally endowed not just with input but also with output
capability, so we use {\it automata} as a cover term for both, including for
weaker `semiautomata' (transition systems, both labeled and unlabeled) that
may lack well-defined initial and accepting states.

While we keep the discussion general, most of our automata will actually have
a single distinguished state (standardly called the {\it initial, resting,} or
{\it reset} state) $q_0 \in Q$, and will be {\it synchronizing} in the sense
that for its transition function $\delta$ there exists some word
$w \in \Sigma^*$ such that $\forall q \in Q ~~ \delta(q,w)=q_0$ i.e. a word that
will take the automaton from whatever current state to the initial state. For
the history of the synchronization property, going back
to \cite{Ashby:1956,Hennie:1964}, and some of the early motivating examples,
see \cite{Volkov:2008}, where one of the oldest unsolved problems in formal
language theory, the \v{C}ern\'{y} conjecture, is discussed.

\smallskip\noindent {\bf Example 1.} {\it The k-wheel $\mathcal{S}_k$}. The
are $k$ states which are hard to individuate beyond the fact that one of them
(not necessarily the initial state) produces output `1' whereas the others are
silent (outputting nothing, implemented as a blank/whitespace symbol `0').
Conceptually, the wheel has no input: we realize this condition by using an
unary input alphabet $\Sigma=\{e\}$ containing the elementary time tick $e$.
The states are arranged cyclically, so that if time flows in an approximately
uniform manner (time ticks arrive approximately uniformly) we will see
$\mathcal{S}_k$ ticking (emitting signal `1') every $k$th tick of the
underlying time signal $e$. Larger wheels, with $10^3 < k < 10^4$ will help
keep approximate synchrony between different {\it timescales}, but we defer
these to Section~\ref{sec:scale}. Here we begin with smaller wheels where some
states may be endowed with self-loops and/or jumps as in the
classic \cite{Bakis:1976} acoustic models (where 3-11 states were typical, and
the loop is unrolled, see $\mathcal{E}_k$ in \ref{ss:ts} below).

\smallskip\noindent {\bf Example 2.} {\it The looped 2-wheel}
$\mathcal{S}_2^1$ has two states $a$ (initial) and $b$ (signaling):

\medskip
\begin{figure}[h]
\xymatrix{a\ar@/^/[r]\ar@(ur,ul) &
b\ar@/^/[l]} \vspace*{-7mm}\hfill
$\begin{bmatrix} 1 & 1\\ 1 & 0 \end{bmatrix}$
\caption{State diagram and transition matrix of $\mathcal{S}_2^1$.}\label{fig:s2}
\end{figure}
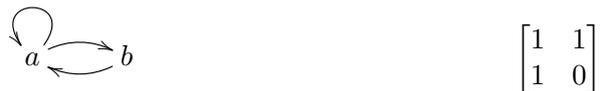

\vspace*{-2mm}\noindent
This simple automaton illustrates an important conceptual difference between
nondeterminism, the free choice between the two paths leaving $a$, and the direct
probabilistic interpretation, where this choice is governed by the outcome of
tossing a fair coin.

\smallskip\noindent {\bf Proposition  1.} As an NDA, the looped 2-wheel
spends, on the average $\frac{1+\sqrt{5}}{3+\sqrt{5}} \sim 0.618 $ of the time
in the looped state, and $\frac{2}{3+\sqrt{5}} \sim 0.382 $ in the other
state.

\smallskip\noindent {\bf Proof} Computing the $n$th power of the transition
matrix shows that after $n$ time ticks there are $f_{n+1}$ paths leading from
$a$ back to itself, but only $f_n$ from $a$ to $b$, where $f_i$ is the $i$th
Fibonacci number. Since $\frac{f_{n+1}}{f_n}$ converges to
$\frac{1+\sqrt{5}}{2}$ the result follows. 

\smallskip\noindent {\bf Proposition  2.} As a probabilistic automaton the looped 2-wheel
spends, on the average, $2/3$ of the time in the looped state, and $1/3$ in
the other state.

\smallskip\noindent {\bf Proof} The transition matrix is $\begin{bmatrix} 1/2
& 1/2\\ 1 & 0 \end{bmatrix}$ and the stable distribution $(\tau, 1-\tau)$ is
the eigenvector corresponding to eigenvalue 1. Direct computation shows that
the unique solution is $(\frac{2}{3},\frac{1}{3})$. 

\smallskip\noindent {\bf Discussion} Unsurprisingly, in both cases the time
spent in the looped state is perceptibly greater than the time spent in the
other state. Perhaps more surprising is the fact that at least in principle we
could make a distinction between the nondeterministic and the probabilistic
model as we take state occupancy observations over longer runs (over $10^3 -
10^4$ elementary time steps) or, what is the same, compute ensemble averages
over many shorter runs.

\smallskip\noindent {\bf Proposition  3.} Given a deterministic $k$-wheel
$\mathcal{S}_k$ (no self-loops or jumps) and output signals $1,2,\ldots,r$
attached to $n_1,n_2,\ldots,n_r$ of the states uniquely so that $\sum_i n_i =
k$, the observed signal coming from the wheel will be $i$ with probability
$n_i/k$.  This obviously also holds if all states have self-loops.

In other words, temporal occupancy statistics endow even deterministic
automata with a probabilistic character. Prop.~3 shows how random variables
can be approximated to arbitrary precision by CMA.

\smallskip\noindent {\bf Example 3.} {\it Wheels within wheels.}  In
$\mathcal{W}$ we assume an outer automaton $\mathcal{S}_2$ and consider its
states $a,b$ to contain wheels $\mathcal{S}_l$ and $\mathcal{S}_m$.  Assume
the inner wheels are sensitive to the same time tick, and produce output at
every $l$ and $m$ steps respectively, with $m, l$ relative primes.  Now, if
this output `1' drives (acts as the time tick for) the outer $\mathcal{S}_2$, we see it
moving across its two states and spending about the same time in each. The
reasoning gets more complicated if $l$ and $m$ are not coprime, but the
average time spent in the two macro-states is 1/2 each independent of the
choice of $l$ and $m \geq 2$

\smallskip\noindent {\bf Discussion} For any finite sample space $\Omega$, 
any probability distribution $P$ over it, and any $\epsilon >0$ we can
construct a wheel that produces output $i$ with the prescribed probability
$p_i$ within $\epsilon$. The fact that state occupancy statistics can be used
as a universal approximator for finite probability distributions justifies the
viewpoint taken in this paper: the {\it random variables} of interest in
linguistics {\it take values} not just in any measurable space but {\it in the
state space of finite automata}.

\begin{figure}[h]
\hspace*{22mm}\xymatrix{r\ar@/^/[r]\ar@(ur,ul) & a\ar@/^/[l]\ar[d]^1\ar@(ur,ul)\\
b\ar[u]\ar@(dr,dl)& t\ar[l]}
\vspace*{4mm}\caption{State diagram of the abstract synapse $\mathcal{R}$}\label{fig:r}
\end{figure}
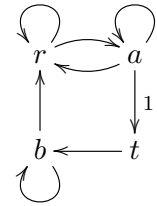

\smallskip\noindent {\bf Example 4.} {\it The abstract synapse $\mathcal{R}$}. We have
$|Q|=4$. The states will be {\it rest, aroused, transmit, blocked} written
$Q=\{r,a,t,b\}$, with transitions arranged cyclically $r \rightarrow
a \rightarrow t \rightarrow b \rightarrow r$. Loop transitions are permitted
everywhere except at the transmitting state $t$, as shown in Fig.~\ref{fig:r}.
There are again just two outputs `0' and `1', so this is effectively
$\mathcal{S}_4^{i}$ where $i$ is 0, 1, 2, or 3 depending on the number of
loops we permit. The `no-loop on $t$' proviso is used for enforcing a well
known constraint that is operative on biological synapses, namely that the
transmission of a signal uses energy and must be followed by some time spent
in the blocked (refractory) state.

\bigskip

\noindent
Actual synapses \href{https://en.wikipedia.org/wiki/Synapse}{come in many
varieties}, and our $\mathcal{R}$ is only broadly inspired by these. What
Fig.~\ref{fig:r} provides is not a fully specified $\mathcal{R}$ but
rather a small parametric family of these. In particular, we have not
discussed whether the synapse can get in aroused state in the absence of
external signals, a matter we shall return to in Section~\ref{sec:ling}.

For more realistic modeling of biological synapses (which come in several
varieties) one would have to provide far more detail on occupancy statistics
of the four states $r,a,t,b$ and specify the underlying timescale e.g. as
centiseconds or milliseconds.  Discussion of the timescales $R_i$ is deferred to
Section~\ref{sec:scale}, here it suffices to say that the $i$ will be integers
between  fixed $\texttt{max}$ and $\texttt{min}$ values: modern physics has
$R_\texttt{min}$ in the attosecond $10^{-18}s$ and $R_\texttt{max}$ in the
exasecond $10^{18}s$ range.

Our Ex.~3 had automata operating on one timescale inside the states of
automata operating on a slower scale. In full generality, we will permit
wheels within wheels within \ldots wheels as long as the outermost automaton
is at $R_\texttt{max}$ or lesser, and the innermost ones are at
$R_\texttt{min}$ or greater. 

\smallskip\noindent {\bf Definition 1.} {\it Clustered Moore Automata (CMA)} are
defined recursively:

\begin{itemize}
\item[(i)] The single-state automaton $\mathcal{S}_1$ with
output alphabet $O = \{0,1\}$ is a CMA
\item[(ii)]  Each CMA can be driven externally by a time-tick $e_i$ where
$i$ is a timescale index ranging between pre-set integer values \texttt{min}
and \texttt{max}
\item[(iii)] If not driven externally, each CMA can be driven internally by
CMA contained in its states. In such cases, the outer CMA's time ticks are
provided by the outputs of the inner CMAs, which much run on a timescale 
strictly faster than that of the outer machine
\item[(iv)] Embedding of machines can stop at any level and must stop
when the fastest timescale $R_{\texttt{min}}$ is reached (which makes further
placement of yet smaller machines no longer feasible).
\end{itemize}

\smallskip\noindent The distinction between being externally and internally
driven is not as sharp as it may appear at first blush. While it may not be
observable what happens at $R_{\texttt{min}}$ or below, Ex.~3 shows how to
construct a balanced binary ticker at $R_{\texttt{min+1}}$, and Ex. 4 shows
how, with access to a ticker, one could inspect the output of $\mathcal{R}$ to
gather a maximum of two bits of information. To communicate more than two bits
asynchronously we use a different CMA.

\smallskip\noindent {\bf Example 5.} The {\it wire} automaton $\mathcal{D}$
will have one state for each symbol in its input/output alphabet plus a rest
state `transmitting nothing'.  From any state, upon receiving symbol $\sigma$,
it transitions to the state indexed by $\sigma$, where it will emit
$\sigma$. (It would be easier to construct wires using Mealy machines, but we
stay with Moore as mixing the two would bring in unnecessary difficulties.)
This enables temporal amplification across timescales: If input is received
on scale $R_i$ output can be produced on any scale $R_j, j \geq i$. 

$\mathcal{D}$ is loosely modeled on the process of object perception whereby
objects cause disturbances, the disturbances radiate outward, and senses are
capable of perceiving them. When machines are embedded in other machines, the
outputs of the inner machine are by design accessible as inputs to the outer
machine, but not necessarily beyond. Wires are a means of instrumenting
internal machines for perception by external machines which may operate on a
considerably slower scale. 

\section{Space and time}\label{sec:scale}

With CMA, nothing comes for free: operations easily taken for granted, such as
writing, reading, or erasing a piece of memory require specific, often
nontrivial constructions. One key issue is {\it persistent storage}, how
information can be preserved over larger timescales. In computers, change of
internal state happens on a scale noticeably faster than the recording of
the result: internal change is on the nanosecond scale, whereas the writing of
hard drives takes several milliseconds and DRAM refresh is still on the
millisecond scale. For the Presocratics, the same difference in timescale must
have been evident in relation to the time it takes to write something down
versus the `lightning speed' of thought.

\subsection{Memory}\label{ss:mem}

We begin by introducing a CMA $\mathcal{T}$ aimed at capturing the behavior of
ordinary Turing Machine {\it tapes} as the behavior of the
CMA.  \citet{Turing:1937} permitted any finite tape alphabet, but this can be
limited to binary 0/1 without loss of generality, and for ease of presentation
we stay with this.  Remarkably, the number of states in $\mathcal{T}$ can be
limited to 4, independent of tape length $l$. Three of these capture the
contents of the tape, and the fourth {\it counter} state will store the
position of the read/write head. The alphabet of $\mathcal{T}$ will contain
not just 0 and 1 but also dedicated symbols $\mu$ for `move left'; $\nu$ `move
right'; $\alpha$ `write 1'; and $\omega$ `erase' (write 0).

The actual contents of the tape will reside in smaller CMA inside the three
states: these smaller machines are hypercubes of size $2^l$ where `being in a
given state at a given time' can be conveniently modeled as the satisfaction
of certain {\it fluents} (time-sensitive predicates) that hold on the 1
coordinates of the point and fail on the 0 coordinates. (These three replicas are
required only for 1-bit error correction -- if this is not a concern the
number of top-level states can be limited to two, see the Appendix.) 

To accommodate the more nuanced multi-scale version of time that we propose
here, the standard event calculus (for extended introductions, see
\citet{Shanahan:1997,Thielscher:2000,Lambalgen:2005}) will have to be
revised: instead of some predicate $F$ obtaining at some time $t$, we will
write $F(i.j)$, where the first index refers to the scale, and the second to
the unit on that scale. The key difference from the standard view is that in
CMA we admit \textit{cyclic} fluents like {\tt Day(3.j)} and {\tt Night(3.j)}.
Since these are non-overlapping events which neither precede nor follow each
other, they fail the axiomatization of precedence and overlap proposed
by \cite{Kamp:1979}. While the urge to ban discontinuous events may be strong,
we would like to reassure the reader that there are no paradoxes, the whole
phenomenon of cyclic fluents is perfectly consistent with the traditional
discrete model of time where we already have interleaving predicates like {\tt
Even(n)} and {\tt Odd(n)}.

Altogether, we will have $l$ fluents $F_k(i,j)$ for each tape position $1 \leq
k \leq l$, with the idea that $F_k(i,j)$ is true iff on scale $i$ at time $j$
position $k$ of the tape has a `1'. Both for reading and writing we need to
know which position is under scan. This information takes up $\log_2{l}$ bits,
and needs to be stored and manipulated just like the data stored on
$\mathcal{T}$. This requires a smaller $\mathcal{T}_1$, which will be stored
in the counter state of the larger $\mathcal{T}$. It follows that staring with a
byte-size automaton $\mathcal{B}$ at the innermost level, we can gain control
of a tape of length 8, which in turn enables gaining control of a tape of
length $2^8$, and so on until we run out of levels at \texttt{max}. Except
for the explicit termination clause, this is reminiscent of the {\it
predicative arithmetic} of \citet{Nelson:1986}. Observe that termination does
not amount to a perceptible limitation on memory, in that with three layers we
are already at $2^{2^8}$ i.e. a tape capable of storing over $10^{154}$ bits,
more than the storage capacity of the entire universe \cite{Lloyd:2002}.

Standardly a TM is composed of a tape with a read/write head and a finite
state control, putting no explicit limits on the complexity of the latter. To
complete our discussion (we defer the details of $\mathcal{B}$ to the
Appendix) we need to consider the limitations for such elementary (control)
CMA. We state the limitations for the top level, but CMA are typically
constructed bottom up, so the following apply for intermediary stages as well:

\vspace*{-3mm}
\begin{itemize}
\item[\bf ss] No single-layer CMA can contain more than a myriad ($m=10^4$) states
\vspace*{-1.5mm}\item[\bf io] No input or output alphabet can contain more than $s=2^8$ symbols
\vspace*{-1.5mm}\item[\bf od] No state can have out-degree $o\geq 8$
\vspace*{-1.5mm}\item[\bf id] No state can have in-degree $i \geq m$ 
\end{itemize}

\noindent
These limits are set generously, more with LLMs than with biological neurons
in mind. In particular, the i/o restriction permits direct modeling of fp8
computations, whereas biological synapses, the inspiration for our
$\mathcal{R}$, appear to be at most two-bit devices (see \citet{Hertz:1991}
10.2). Using one $\mathcal{T}$ as control, connected to a tape by a wire,
requires only four symbols $\mu,\nu,\alpha,\omega$. By cascading, larger
out-degrees could easily be modeled as soon as $o=2$ is permitted: having
larger numbers serves no purpose beyond avoiding too many layers of cascading.

\smallskip\noindent {\bf Definition 2.} We define any temporal {\it instant j}  as
being located on some fixed timescale $R_i$. Time-sensitive predicates ({\it
fluents}) take $(i.j)$ instances as arguments.  We write
$\mathcal{A} \in \mathcal{B}$ if the automaton $\mathcal{A}$ is inside a
single state of $\mathcal{B}$.

\smallskip\noindent The `element of' relation $\in$ is eternal in the sense
that all instances $j$ on timescale $i-1$ belong in a single time tick $k$ on
scale $i$. This means that we can reuse the $\in$ symbol for time unit
containment and write $\forall j (i-1.j) \in (i.k)$. What is not necessarily
eternal is the structure of automata which may, on larger timescales, actually
lose or gain states and transitions.

\smallskip\noindent {\bf Example 5.} Budapest phone numbers. The automaton
(regexp) describing phone numbers evolved, on a decade scale, as follows.
1960s: 6-digit numbers starting with 1 or 3; 1970s: other leading digits
gradually added; 1980s: switch to 7 digit numbers, initially adding a leading
digit 9; 1990s: other leading digits gradually added; 2000s: city code 1 added
as obligatory first digit; 2010s: inter-urban prefix 06 or international prefix
+36 made obligatory. What started out as 123456 eventually became +3619123456.
We leave it to the reader to arrange the small automata corresponding to each
regexp within successive states of the larger automaton moving in decade steps. 

\subsection{Temporal structure}\label{ss:ts}

At first blush it looks as if embedding CMA in one another will create a
tremendous variety of temporalities.  Our first theorem establishes that this
is not so: there are only five families of discrete temporal structures up to
bisimulation, and most of them are rather similar to the ordinary concept of
discrete time that is widely used in computer science, and enforced by clock
signals in computer engineering.

We begin by exemplifying these five kinds of structures Z, N, P, L, and C. The first
three are `infinitistic' in the sense that CMA bisimulation cannot distinguish
them from the well-known infinite structures $\mathbb{Z}, \mathbb{N}$ and
$-\mathbb{N}$, but for ease of presentation we begin with the finitistic ones.

C stands for {\it cyclic}, and our first example, the $k$-wheel
$\mathcal{S}_k$ shows how this can be realized for any $k \leq m = 10^4$. This
already gives a myriad examples of non-congruent cyclic temporal structures,
but this can be pushed much farther. Even if we stay within the per-level
limitation $m$ we are imposing on elementary structures, we could build a much
bigger system. Given $r$ wheels
$\mathcal{S}_{l_1}, \mathcal{S}_{l_2}, \ldots \mathcal{S}_{l_r}$ arranged as
the inner wheels of an outer wheel $\mathcal{S}_r$, the Chinese Remainder
Theorem says that the entire structure will return to its initial state in
exactly lcd($l_1,l_2,\ldots,l_r$) steps.

There are 1229 primes less than a myriad, so we start with an outer loop of
1229 steps, and place wheels with length 8192, 6561, 3125, 2401, 1331,
2197, \ldots (maximum prime power length $p^{m_p} <m $) in these. Such a
system would in principle take over $10^{4348}$ steps to return to its initial
state. This number is vastly larger than the entire gamut from $10^{-44}$
(Planck time) to $10^{20}$ (lifetime of the universe) countenanced by
contemporary physics. 
The naive (Presocratic) theory that is already
sufficient for the linguistic applications discussed in Section~\ref{sec:ling}
has considerably smaller gamut, $10^{14}$ instead of $10^{64}$, but the point
is the same: `cycles' of such extraordinary length cannot be distinguished
from the infinite system Z indexed by the integers in $\mathbb{Z}$. 

L stands for {\it limited}, and the structures $\mathcal{E}_k$ are obtained by
removing from the wheel $\mathcal{S}_k$ the transition leading to the initial
state. This way we obtain a chain of $k$ steps. This again yields a myriad
non-congruent (to each other or to anything from C) linear structures. L
structures, finite intervals with a definite beginning and end, are implicit
in contemporary Big Bang/Big Crunch models and quite explicit in both Old and
New Testament eschatology: the world has a definite beginning and a definite
end.

In Z, which is modeled on $\mathbb{Z}$, we start numbering by 0
from \texttt{now}, using negative integers for predecessors (past time
units, \texttt{before}) and positive integers for successors (future time
units, \texttt{after}) as expected. In other words, we rely on `event
time' \citep{Reichenbach:1947} as our reference frame. Note that the modern
Unix time, based on 64-bit rather than 32-bit counters, is capable of handling
292 billion years in either direction. Since this exceeds the estimated past
and expected future lifetime of the universe by more than an order of
magnitude, $C_{2^{64}}$ temporality is effectively indistinguishable from
Z-time.

N is analogous to $\mathbb{N}$, with a definite starting point, but no
endpoint. The future works as expected, but the past is limited to time-ticks
since the origin. Unix time with unsigned quantities is a possible example,
as are temporalities associated to imperfect linguistic expressions. 

The final option P is the mirror image of N, with a definite endpoint but no
starting point -- the most common usage is countdowns, but this is the
internal timeframe of telic expressions as well. Modern cosmology
countenances N-models (those that have a Big Bang but no Big Crunch) and could
perhaps permit P-models as well.

Common to all five notions of time is the existence of a definite instance
preceding, and a definite instance following \texttt{now}, or any other
instance. It is this Peano-style succession (no two units are succeeded by the
same unit, and only one unit succeeds any unit) that distinguishes the
Presocratic concept of time from the modern, continuous notion of time, where
`immediately preceding' and `immediately following' make no sense (though the
disjointness of the instances is preserved). Presocratic thought is not very
specific on which of these five models is employed on any given scale, and for
N- P- and L-time leaves open the question of whether the initial and/or final
units are to be included or not.

Though simulation in one direction is often possible, e.g. a 4-wheel can
simulate a 2-wheel, bisimulation is not: the inventory listed above is all
that elementary CMAs are capable of.

\smallskip\noindent {\bf Theorem 1.} {\it Single-scale structure} There exist
five (families of) structures Z, N, P, L, and C such that any timescale
associated to a CMA has a bisimulation with one of these.

\smallskip\noindent The proof, not presented in detail, is easy but tedious,
requiring the checking of all combinations of the canonical temporal
structures. For example, the direct product of a C and an L structure
(effectively seen by embedding them as subautomata in $\mathcal{S}_2$) will
again be C, if we permit a self-loop at the end of the L subautomaton, or L,
if we don't. Note, however, that the bisimulations we consider are {\it weak}
in the presence of C structures in that `which happened first' may not have a
meaningful answer for events on cyclic scales. $\square$.

\section{Linguistic applications}\label{sec:ling}

CMA timescales are neutral between the modern view described
in \cite{tHooft:2014} and the `naive' view characteristic of Greek
philosophers from the Presocratics to the Hellenistic age. Since the
modern view based on powers of 10 (scientific notation) will be thoroughly
familiar to the reader,
in \ref{ss:naive} we concentrate on reconstructing the naive worldview, and
pointing out salient differences from the modern view.

In \ref{ss:bm} we use CMA to get a handle of some primitive categories of
linguistic description, typically encoded in {\it bound
morphemes}. Unsurprisingly, timescales easily lend themselves to describing
tense/aspect markers, and perhaps more surprisingly, to the description of
a naive theory of causation as well. 

Finally, in \ref{ss:gb} we use CMA to implement grammatical systems capable of
island parsing and disambiguation.

\subsection{The naive worldview}\label{ss:naive}

It is this, somewhat crude, view that underlies the use of ordinary language,
and the reliance on {\it instants}, point-like elementary time particles
without perceptible duration, is common to all formal ontologies attempting to
characterize natural language, including the influential \cite{Hobbs:2004}.

First, we need to pick a time unit in a manner that will make the events we
are most concerned with easily expressible. Science settled on the {\it
second} (s) but for the naive view we will measure time in {\it hearbeats},
obviously a more variable (and less reliable) unit. Either way, there is a
unit that anchors the \textit{real time} scale $R_0$.

Going downwards, we find the \textit{centisecond} (cs) timescale appropriate
for fast events. In general, events not separated by 2-3 cs cannot be told
apart by unaided perception, and structured human activity, such as speech,
also requires units at least this long. (Normal speech sounds take 7-10 cs,
but can be only 2-3 cs in extreme fast speech.) In vision, a still frame
lasting 3-4 cs is sufficient for making movies appear natural, and using
stills separated by 1cs fully eliminates film artifacts like strobing,
flicker, and motion blur. Human reaction times are on the order of 10-30
cs. Altogether, events need to last several cs for discrimination, and even
subliminal perception is hampered under 3cs and undetectable under 1
cs \citep{Ionescu:2016}.

We could call $R_{-1}$ the `blink of an eye' scale (an eyeblink takes about 10
cs) but we emphasize that the unit on this scale, what we will call an {\it
instant}, is soft in both directions. Even though microsecond measurement
technology is available off the shelf, it makes no sense to measure eyeblinks
down to this level. Similarly, it is not a fixed number that describes the
conversion factor from instances to heartbeats but rather a probability
distribution with noticeable variance.

Importantly, the centisecond instant constituted an absolute barrier of
observability for the Presocratics the same way as Planck time units
constitute an absolute barrier for physics today. Current measurement
technology is in the zeptosecond range, still separated from Planck time by
over 23 orders of magnitude, so assuming an absolute observability barrier
$R_{\texttt{min}}$ is just as justified today as it was in antiquity. Progress
in metrology pushes the minimum detectable unit down by half an order of
magnitude every decade, but does not remove the conceptual limitation that we
are incapable of detecting phenomena below some limit.

Moving upward, at the \textit{kilosecond} (ks, quarter hour) timescale slow
motion, such as that of the sun in the sky, is barely perceptible, though on
the \textit{day} scale, which contains 96 quarter-hour (86.4 ks) time slots,
such motion is evident. The ks scale is well suited for describing changes in
human perceptions regarding both internal states (hunger, sleepiness,
tiredness) and external states (temperature, light, weather), for which the
second scale, let alone the centisecond scale, are too detailed. It takes
another two rough orders of magnitude (ROoM) for the change
of \textit{seasons} to be clearly discernible, and the largest celestial
cycles of which knowledge can be safely ascribed to the Greeks of antiquity,
the Metonic cycle (235 months) and Saros (223 months), require another
two-three ROoMs. These comprise the \textit{generation} scale both in terms of
human development stages from child to adult to elder and in terms of
successive generations. Finally, we have the \textit{aeon}, a time unit of a
few thousand years.

CMA can be used equally well to describe the modern view: for this we can
simply use 10-wheels within 10-wheels within \ldots 10-wheels going down from
$R_{\texttt{max}}$ to $R_{\texttt{min}}$, setting the former at the expected
lifetime of the universe (currently estimated below $10^{18}$s).  The naive
view requires larger, and unevenly sized, wheels, on the order of $10^2 -
10^3$. The conversion factors $k$ are not specified exactly, but remain under
an equally vaguely defined constant $m$, the {\it myriad} of Greek, Latin, and
Classical Chinese, which we take to be $\leq 10^4$.  In terms of temporal
adjacency, we have two relations to consider: the same instant on different
scales, and different instants on the same scale, and our notation in
Definition 2 reflects this.

The naive worldview incorporates cyclic timescales. For the sake of simplicity
assume the underlying time to be $C_2$, say day/night, and assume a worker who
can perform some task (dig a trench, add lines to a new copy of some
codex, \ldots) every day. If the worker can overwrite a piece of memory that
has persistence on a larger scale, \textit{nulla dies sine linea} is not in
vain: days and nights come and go, but the trench is ever longer, the codex is
getting copied. (It would be ahistorical in the naive context, but the
principle is not any different for our Turing machine tape $\mathcal{T}$.)

What appears to be endless (repetitive) toil on the day/night scale will
correspond to significant progress on the next larger scale. But if there is
no external (higher scale) record keeping, the automaton must preserve all
information in its own state space, and this is limited to less than 14 bits
by {\bf ss}. If the automaton eventually halts, it needs to be externally
reset for further use, but the overall time structure provided by it is still
cyclic if these resets (synchronizing) are taken into consideration as argued
by \citet{McCulloch:1945}. 

What do we mean when we say that a fluent $p(i.j)$
holds at some time unit $j$ on timescale $i$?  Looking at this on the next
smaller scale $i-1$ there are two plausible definitions: $\forall (i-1.k) \in
(i.j)\; p(i-1.k)$ (the $\forall$ definition), or $\exists (i-1.k) \in (i.j)\;
p(i-1.k)$ (the $\exists$ definition). The choice becomes irrelevant only on the
fastest timescale $R_{\texttt{min}}$ that is accessible to measurement, which
we set as the $R_{-1}$ (cs) scale for the Presocratics. On ordinary scales,
such as $R_1$ (ks) we either countenance the possibility of inhomogeneity,
that Hermocrates can be eating for a quarter hour without eating at every
heartbeat, or we insist on the $\forall$ definition, which forces homogeneity
not just one scale down, but iteratively down to all accessible scales and
perhaps beyond. An important consequence of the $\forall$ choice is what is
known in phonology as the \textit{Absolute Slicing Hypothesis} (ASH)
\citep{Goldsmith:1990}, that boundaries of units on scale $i$ must be
coterminous with boundaries of edge units on the next smaller scale.

This hypothesis is hardwired in the modern view, which relies entirely on the
standard real line $\mathbb{R}$, making the division into ever smaller
timescales containing exactly 10 units for one above it (or $10^3$ as embodied
in the SI system of prefixes from quetta to quecto) something of a historical
accident, the widespread use of base 10 counting. The {\it essence} of such a
system is continuous, the smaller scales appear only for practical reason and
carry no theoretical significance. In contrast, the naive view is essentially
discrete, with continuity akin to that of the nonstandard reals, ordinary
reals enhanced with nilsquare infinitesimal \citep{Bell:2008} instances. 

Historically, phonology has moved away from the ASH as there is clear evidence that
independent articulators (such as the vocal folds, the tongue body, and the
lips) do not move in perfect synchrony, so that the articulation of a single
sound, such as a /p/ in /apa/ may begin with lip closure preceding the
cessation of voicing, or the other way around, may start with lip closure only
after the opening of the vocal folds (for a cs-level analysis of voice onset
time effects, see \cite{Cho:1999}). Phonology provides a clear contemporary
example where using the $\exists$ definition is more advantageous, but the
fact that events following their own rhythm need not change at the same
instance must have already been evident to the Presocratics, who knew
perfectly well that the moon can appear during the day. But if satisfaction on
all smaller scales is not required for the truth of a fluent $p(i.j)$, what is
required?

\smallskip\noindent {\bf Definition 3.} $p(i.j)$ holds iff we have a 
{\it contiguous and preponderant} stretch of subunits $(i-1.b),
(i-1.b+1), \ldots (i-1.b+l)$ at which $p$ holds.

\smallskip\noindent However many $R_{i-1}$ units there are in a $R_i$ unit,
it simply cannot be the case that both $p$ and $\neg p$ units are
preponderant, so the definition excludes $i$-level units that harbor
contradictions. The possibility that $p(i.j)$ is undefined is left open, and
is seen in practice for all cases where $p(i-1.k)$ is \textit{alternating}:
days and nights, breathe in, breathe out. For these, assigning truth values
{\it During a quarter hour, are people breathing in or breathing out} makes no
sense. 

What happens if 50.1\% of $p$ stands in opposition to 49.9\% of $\neg p$?
Modern theories of knowledge representation would have to say that we have a
preponderance of $p$, but this is very far from Presocratic thinking, where
preponderance requires a perceptible majority, maybe as much as two
thirds. Units where neither $p$ nor $\neg p$ are dominant are simply
\textit{transition} units, perhaps to be analyzed on the next smaller scale,
if that is still accessible. Note that on the appropriate scale contiguity
will make the $p$ or $\neg p$ question clearly settled for most lower level
units, except possibly the one where the transition is taking place. 

For purposes of representing semantic knowledge, we use \textit{frames} as
in \texttt{FrameNet} \citep{Baker:1998,Fillmore:1998,Ruppenhofer:2006}, and
temporally linked series of frames called \textit{scripts} or
(skeletal) \textit{stories} \citep{Schank:1995}.  We will also use cognitive
structures that are not quite Sausserian signs, in that it is only the idea
that is fully accessible to us, the form associated to the idea is less
pronounced -- we will call these \textit{schemes} or \textit{schemas}. We
begin with a well-known example, the \texttt{Exchange} schema, used in the
analysis of verbs like \textit{buy} and \textit{sell} \citep{Hovav:2008}.

\smallskip\noindent {\bf Example 6.} \texttt{Exchange}. The 
temporal aspects are captured in $\mathcal{E}_3$, a 3-state linear machine
that requires no input, being driven by the time tick alone.  We call the
initial state $b$ ({\tt before}) and the final state $a$ ({\tt after}). There
is just one fluent, \texttt{x has y}, that matters:
we have \texttt{(seller has goods)(b), (buyer has money)(b), (seller has
money)(a)} and \texttt{(buyer has goods)(a)}.

\smallskip\noindent This automaton operates on the
heartbeat scale, and we could go in more minute detail, because these days the
standard commercial exchange involves a great deal of background machinery:
the buyer has a credit card, or better yet, a cellphone that acts as one, the
seller has a credit card terminal, the buyer and the seller both have bank
accounts linked to these, and the exchange of money is effected, perhaps
substantially later, by some protocol neither buyer nor seller are fully in
control of \citep{Brandt:2011}. 

Note that the value of the fluent \texttt{has} is underspecified for
\texttt{now}. It simply makes no difference whether we postulate that at the
moment of the exchange both the seller and the buyer have the goods or
neither, or whether we conceptualize it as the seller giving up possession of
the goods before or after getting possession of the money. These details can
matter for the purpose of debugging a transaction, but even there the
governing considerations will not be framed in terms of such microanalysis but
rather in terms of which fluent failed \texttt{before} or \texttt{after}. For
example, if the transaction went through, and \texttt{buyer has goods}, the
seller should be able to obtain the money from the credit card company even if
the buyer dies before his credit card is paid off. How the card company gets
the money from the estate of the buyer is not the seller's concern, and the
process may play out on the season scale, what happened on the hearbeat scale
is irrelevant for this.

For human action that is driven by social conventions, such as buying and
selling, the naive model still rules supreme, highlighting all and only the
factors that matter. But for action driven by natural law, the difference
between the contemporary and the naive view could not be more stark. We
illustrate this on the example of gravity, which we can posit in the Newtonian
form as $F=gmM/r^2$.  Einstein field equations would be even farther from the
naive theory, but this is already sufficient for making the point.

\smallskip\noindent {\bf Example 7.} \texttt{Gravity}. This is a two-state 
cyclic automaton with one (start and end) state being the Aristotelian resting
state, where some form of support counterbalances gravitational pull, and in
the other state no support is present and the object begins to fall. 

\smallskip\noindent Naive as this theory may be, it still lets us draw some
key conclusions: that things, unless supported, will fall; that every fall
ends in a (synchronizing) resting state; that the stars must be supported by
some celestial sphere; and so on. Our contemporary selves may find these
conclusions ridiculous, but nearly two thousand years after the Presocratics,
in the 14th century Jean Buridan still maintained a theory of circular impetus
to explain celestial mechanics, and it will be another three hundred years
until Newton manages to unify the theory of sublunar and superlunar realms. To
this day, children under six will regularly believe that if we spin in a
circle a rock attached to the center by a string, if we abruptly cut the
string the rock will continue on a circular or spiral orbit. Many retain this
naive view into adulthood.



\subsection{Bound morphemes}\label{ss:bm}

There is a great deal of evidence in favor of a CMA model of temporality in
natural language, coming from a variety of bound morphemes that mark tense,
aspect, frequency, duration, and related distinctions in a variety of ways in
most, if not all, languages spoken around the world. We will begin with tense
and Aksionsart, but we cannot possibly do justice to the broad linguistic and
philosophical tradition here: for a survery sympathetic to our point of view
consult \citep{Fernando:2015}.  The evidence for different scales is manifest
in may languages like Tamil:

\begin{table}[h]
    \centering
\hspace*{-3mm}
    \begin{tabular}{lll}
        \textbf{Past} & \textbf{Form} & \textbf{Gloss} \\
        \hline
        Immediate &  vantu vi\d{t}\d{t}\={a}n & He just came\\
        Recent  &  vantirukki\d{r}\={a}n & He has recently come\\
        Remote  &  vant\={a}n & He came\\
        Historical & vanti\d{r}unt\={a}n & He had come long ago\\
    \end{tabular}
    \caption{Distinctions in Tamil past tense (3sg masculine)}
    \label{tab:tamil_past}
\end{table}

\vspace*{-3mm}\noindent Immediate past is on a heartbeat scale, {\it he came a moment ago}; recent
past on the quarter hour scale {\it he came earlier today}; remote past is on
the day scale {\it he came yesterday/a few days ago}; historical past is on
the generation/aeon scale {\it he came in ancient times}. Taking the moment of
utterance to be 0, he arrived at some instant {\it (i.-k)}, where $k$ is a small
number `a few', and $i=0,1,2,4$ respectively. We also mention Tagalog which, in
addition to Immediate, Recent, Distant, and Remote Past, also has a
Mythological Past used for legends and ancient history.

Analogous distinctions in the future exist both for Tagalog and for many other
languages for Immediate future (happening very soon); Near future (expected
within a short time); Distant future (further away but still definite); 
Uncertain/hypothetical future (may or may not happen): here again we refer to
some instant $(i.l)$ with $l$ a small positive number and $i=0,1,2,4/5$.

Besides Tagalog and Tamil, etymologically unrelated languages like Kalaallisut
(Greenlandic Inuit), Yagua, some Bantu languages, and perhaps many others (the
World Atlas of Language Structures \citep{Dryer:2013} suggests the majority)
will make more than one scalar distinction in the past, the future, or both. To
be sure, not all of these distinctions are purely morphological (various
multi-word constructions are common) but the phenomenon is undeniable for
tenses and for aspect markers to which we turn now.

Lexical aspect (Aktionsart, see \citet{Vendler:1967}) requires only small
automata. Durative static actions, better named {\bf states} like {\it know},
require a single state with a self-loop, $\mathcal{S}_1$. Once in that state,
on the relevant timescale there is no way out -- perhaps you can unlearn on a
longer scale, but you can't unknow. An even simpler machine, one state and no
self-loop, is appropriate for punctual {\bf semelfactives} like {\it
blink}. Such actions are assigned a single time unit on the scale below the
narrative event line, typically $R_{-1}$. {\bf Achievements} like {\it find}
are two-state automata $\mathcal{E}_2$ with no self-loops, as opposed to {\bf
accomplishments} which have a self-loop on the first state. Finally' Vendler's
{\bf activities} like {\it walk} have self-loops on both states.

In general, perfect forms indicate that the underlying temporality is
terminating, what we called P-time in 2.2.  Nearly half of the languages
considered in \citep{wals-68} have some kind of perfective, often formed by
some auxiliary word {\it has, finish, already}. Similar considerations could
help organize imperfective/pluperfect forms and, with the aid of cyclic
temporalities, habitual/frequentative ones as well, but this would take us far
beyond the bounds of this paper. Instead of investigating the many ways
temporal pointers {\it can} be updated by syntax, we will concentrate on the
one case where they {\it must} be updated.

For this, we use the relation {\tt a cause b} where, under any theory of
causation, the cause must precede the effect, some kind of temporal update
between {\tt a} and {\tt b} is necessary. In general, the naive worldview
assumes inertia: changes are caused by some \textit{force} and conversely, if
no forces are present, things will continue as they are \citep{Talmy:1988}.
Consider the tale of the Crow and the Serpent from
the \textit{Pa\={n}catantra} (Bk. I/5).

\begin{quote}
Once upon a time there lived a crow couple, who had built a nest on the top of
a tree. But unfortunately the tree was inhabited by a serpent at its
bottom. So the serpent used to crawl up the tree and eat all the eggs that the
lady crow used to lay. The crow couple were deeply grieved and when this
happened time after time, then they decided that the serpent was to be get rid
of by a plan.
  
So the crow then approached his friend the jackal and asked for a plan. The
jackal then told him to go and fetch a costly thing from the palace of the
king and throw the thing in the burrow of the snake. The crow went to the
palace, and stole a necklace of the queen while she was bathing. The guards of
the palace ran after it. The crow then dropped the necklace in the burrow of
the snake beneath the tree.

The guards on reaching the bottom of the tree, found the necklace guarded by
the serpent. Then they killed the serpent and recovered the necklace. So the
crow family was now happy that their eggs were safe now. (Translation
\href{https://nriol.com/indianparents/indian-tales/crow-serpent-story.asp}{source})
\end{quote}

\noindent
The causal chain proceeds from 1: \texttt{(lady crow laying eggs) enable
(serpent eating eggs)} which in turn
2: \texttt{1 cause (crow grieving)};\\
3: \texttt{2 cause (crow asking for plan)};\\
4: \texttt{3 cause (jackal providing plan)};\\
5: \texttt{4 cause (crow follow plan)};\\
6: \texttt{5 cause (crow steal necklace)};\\
7: \texttt{5 cause (crow drop necklace in burrow)};\\
8: \texttt{7 cause (necklace in burrow)};\\
9: \texttt{6 cause (servants want necklace back)};\\
10: \texttt{7 cause (servants go to burrow)};\\
11: \texttt{8 cause (servants kill snake)};\\
12: \texttt{11 cause (serpent dead)};\\
13: \texttt{12 cause (crow family happy)};

\smallskip\noindent Under 1 we have {\tt enable} rather than {\tt cause},
since the snake could, at least in principle, refrain from eating the eggs.
Otherwise, the events are causing one another just by taking place in the
right order, and that physical and mental events/actions are freely
intermingled in the causal chain. Compound events can be built up by compound
actions, e.g.  the crow needs both to steal the necklace \textit{and} to drop
it down the burrow to trigger the servants' action, but the overall strength
of the chain is not affected by this. The habitual aspect at the beginning of
the tale is evident: it is not simply the one-time eating of the eggs that
spurs the crows into action, but that this happens ``time after time''. (As
mentioned above, wherever such aspects are present, time is not fully linear,
loops are involved. While \cite{Fernando:2022} does not consider cyclic
fluents, the kind of action logic he proposes is clearly related to the one
suggested here, treating actions as transitions between the states of a finite
automaton, and clustered Moore automata can perhaps be considered an extension
of his interpretation.) 

More formally, the above sketch of the clearly causal steps relies on a
sequence of time units $t_2,\ldots,t_{12}$ not necessarily adjacent to one
another, but all on the $R_1$ time scale (so the first index of our double
indexing scheme could be dropped here). Let us begin at $t_2$
where we have \texttt{(serpent eating eggs)} \texttt{cause} \texttt{(crow
  grieving)} abbreviates a specific substitution instance of a more general
scheme (\texttt{x parentOf y}) $\wedge$ (\texttt{y die}) \texttt{cause}
(\texttt{x grieve}). This scheme, `death of offspring causes parental grief'
is clearly causal in the weak sense we are considering: death of offspring
will indeed be followed by parental grief.\footnote{This is the default
  expectation: there may be some overriding factors that block the inference,
  but this is quite typical of natural language inference.}

Similarly, going from $t_2$ to $t_3$ relies on general schemes, e.g. that
\texttt{grief} \texttt{isA} \texttt{pain}, that sentient beings attempt to
avoid pain, that thinking beings make plans not to repeat situations where
they get in pain, and that asking a friend for help is a reasonable course of
action. It also relies on knowledge specific to the situation such that
\texttt{jackal} \texttt{friendOf} \texttt{crow}. Later steps of the analysis
follow pretty much along the same lines, assuming little more than
conventional pieces of lexical knowledge, e.g. that friends try to help
friends (steps 4 and 5), what it means to put plans in action (steps 6 and 7),
how obtaining something from the dwelling of a dangerous animal requires
killing it (step 11), etc. Typically these schemas are accessible as
extralogical axioms or `meaning postulates' \citep{Zimmermann:1999} attached
directly to lexical items, and the task of the logic is to spell out the ones
not stored in the lexicon and the pattern recognition mechanisms required for
instantiating the schemes.

\subsection{Grammar building}\label{ss:gb}

At the beginning of Section~\ref{sec:scale} we already emphasized that in the finite
state realm nothing comes for free: operations easily taken for granted by
grammar writers, such as rewrite rules, movement transformation, pattern
matching, variable substitution, quantification, lexical lookup, or `merge'
all require nontrivial constructions. In the Appendix we describe how to build
limited Turing machines from CMA, and at a theoretical level this takes care
of all devices required for grammar writing. But as a practical matter, we aim
at a grammar architecture where string manipulation is pretty much restricted
to {\it linearization,} at the last stage of a derivation.

In contrast to the (transformational) generative grammar tradition, where
lexical lookup is the last stage, here we follow the categorial grammar
tradition and begin with the lexical entries, morphemes, words, or multi-word
constructions that get activated by the phonological component. We assume
that each of these is a graph composed of a few nodes, with edges to and from
some further graphs, limited by the in-degree and out-degree restrictions {\bf
id} and {\bf od}). We use CMA to model the activation behavor of such graphs,
beginning with the synapse $\mathcal{R}$ of Ex.~4. 

Distinguishing the synchronizing {\it rest} state $r$ from the {\it aroused}
state $a$ permits a primitive form of counting, whereby it takes not just one
but two `1' impulses for $\mathcal{R}$ to move to the transmit stage $t$. By
manipulating the size of the main loop and the presence/absence of self-loops
we could make the automaton count up to larger threshold values, but two will
be sufficient for our purposes of modeling {\it spreading
activation} \citep{Quillian:1968,Collins:1975} in the network composed of
lexical items and naive laws. 

Our two-input synapses are idealizations, corresponding to information
transmission across larger units built from many, possibly thousands or even
millions of biological neurons. Consider the naive law `death of offspring
causes parental grief'. This is not a culture- or language-specific regularity
of the kind Montague Grammar encodes in knowledge postulates. It is universal
among humans and, as a matter of fact, clearly already in effect among higher
animals, apes and cetaceans in particular. We don't need differential
equations or other highly sophisticated techniques for using it in predicting
behavior, again the naive theory contains all and only the information that
matters. We need some static links {\tt x parentOf y} and some background
knowledge that {\tt person has emotionalState} and {\tt grief isA
emotinalState} before we can cast the law in terms of an active link {\tt person
has grief}, but we emphasize that the mechanism is pre-linguistic: grief will
be triggered by the person seeing their offspring die, finding this out by
linguistic means (someone telling them) is not necessary.

Once the configuration {\tt x parentOf y, x know (die y)} is active, the
(naive, empirical) law says that {\tt x has grief} will be active. There is
clearly an associative link from death to grief, and if the former is
activated, the latter will automatically move from $r$ to $a$ or, if it was
already in $a$, from $a$ to $t$. The other wave of activation comes from the
subject {\tt y} of death, who is already linked to the parent {\tt x}, so we
can reasonably assume that {\tt x} will grieve, as the (prelinguistic) grief
node that {\tt x} has will now be doubly activated, via {\tt y} and via {\tt
death}. Notice how this simple mechanism already avoids the age-old
philosophical puzzles (suppose that {\tt y} is the son of {\tt x} but {\tt x}
does not know this, her son dies, how come she is not grieving?) that go back
to the Electra paradox of Eubulides.

We will not discuss here how we can use CMA to perform the pattern recognition
(phonology) required for finding the lexical items in the speech stream,
suffice to say that fp8 finite automata perform increasingly on a par with
humans \citep{Radford:2022}. The morphology is already well covered by finite
state transducers \citep{Koskenniemi:1983a}. This model that has been extended
to high-precision constraint-based descriptive grammars, but we select here a
perhaps more intuitive grammar formalism, {\it island
parsing} \citep{Carroll:1983}, as it is a better fit with highly lexicalized
theories that operate with complex verbal entries such as FrameNet.

Suppose therefore that the phonology already identified the morphemes in
{\it Eleanor broke the record}, and that the morphology analyzed {\it broke}
as {\it break.PAST}. The syntax requires reference to abstract grammatical
categories, such that {\it Eleanor} is an NP, {\it break} is a noun or a
transitive verb, {\it the} is an article, and {\it record is a noun, transitive
verb, or adjective}. It further requires storage of abstract patterns,
e.g. that Art followed by N may be an NP. As long as we have nodes in the graph
for abstract categories (and we must, the data compression they facilitate is
irresistible) we need only simple automata such as an instance of
$\mathcal{E}_3$ which, upon receiving a signal from the (phonologically
activated) Art automaton moves from its initial resting state to its next
state where, upon receiving an N signal it will move to its final state and
emit an NP signal.

Note that word {\it record,} or more precisely, the lexical entry activated by
this phonological string, actually emits three signals N, Vt, A in parallel,
but only one of these fits the (Art N) NP pattern. As there are no (Art V) and
(Art A) patterns in English, the readings associated with verbal and
adjectival {\it record} simply do not surface: we built the NP island but not
the VP or AP. For the noun, there is still ambiguity between {\it record$_1$}
`vinyl or bakelite disc containing music'; {\it record$_2$} `database record';
and {\it record$_3$} `sports record'.

The process repeats with the (Vt NP), a VP pattern activated by the newly
formed NP and the verbal reading of {\it break} (as opposed to the nominal
`intermission', which again gets no reinforcement from any pattern and dies
out). Finally, we complete the NP VP sentential pattern, and may be capable of
bringing information about Eleanor to bear on the ambiguity. If she is an
athlete, we expect the last reading; if she is a famous hacker, we expect the
second; and if she is known to be clumsy with things, we expect the first. Not
knowing anything about her we are forced to keep all three readings active and
hope that larger discourse context will later disambiguate. 

\section{Conclusions}\label{sec:conc}

We have introduced a definitional variant of the classic Moore automaton
(transducer), but without taking seriously the notion of timescales that comes
with it, clustered automata would be just regular Moore automata. While we
have no doubt about the psychological reality of grammar, we don't
currently expect micro-electrode arrays to be capable of targeting conceptual
representations/lexical entries directly. Therefore, the best research path
forward is locating analogous structures in
LLMs \citep{Templeton:2024,Dao:2024}.

Theorem~1 leaves open the cosmological problem of whether the temporality of
our Universe follows Z, N, P, L, or C. By the construction of CMA, workers
cannot guarantee a linear timescale either above or below themselves. Calling
modern physics to our aid is of no help here, since Poincar\'e recurrence time
for the universe is on the order of $e^{10^{120}}$ Planck units, which is so
far beyond the $R_5$ scale on which our civilization is located as to render
speculation on this matter meaningless.


\bibliography{ml}
%


\section*{Appendix}

\noindent
{\bf 4.1 $\mathcal{B}$} Because we limited the tape alphabet to 0/1, storing a single byte already
requires 8 cells on the tape. The resting state of $\mathcal{B}$ is one where
the read/right head is on the 0th of these cells (the low bit), and the
contents of these bit-size cells are what they are. To store any byte on
timescale $R_i$ we need to assure that elementary time steps $(i.1)$ leave
these cells (conventionally organized in an 8-dim hypercube) undisturbed for a
myriad steps, as long as there are no read/write signals present {\it and} to
assure that the read-write pointer is synchronizing (decaying to the low bit).
For both tasks we use CMA operating on scale $R_{i-1}$: the head position is
stored on a chain $\mathcal{E}_8$ where there are only transitions going down,
while the byte is stored in the hypercube where each node has only self-loops.
As long as there are no other signals, the head pointer will decay to 0 in
(i-1.8) whereas the stored byte persists. The symbols $\mu$ and $\nu$ move the
head in the expected manner, and the symbols $\alpha$ and $\omega$ move state
at the coordinate pointed at by the head. The entire transition table would
take over 10k cells ($256\cdot 8 \cdot 5$) to spell out, but the point is
clear: we can build richer memory cells as needed.

\smallskip\noindent{\bf 4.2 Building $\mathcal{T}_1$ from $\mathcal{B}$}
Now that we have our hand on a memory cell capable of storing 256 different
values, we can repeat what we did with $\mathcal{B}$ but this will be a bit slower:
we need to make this pointer decay in $(i+1,k)$ steps for some small $k$. The
fastest way to do this is coordinatewise, so $k=8$, but this takes some care
in designing the transition table and, more important, is not repeatable for
the next higher level up, where we would need to make $2^{256}$ coordinates
decay. For practical purposes this does not matter: after all, we know how to
build large memory from elementary components that can be updated on a
nanosecond scale, and can be made to persist on a generation scale at I/O
speeds. But we have no means for persisting anything beyond that: less than
$2^7$ generations have passed and we still live in the same aeon as the
Presocratics, yet we can know their thoughts only in a fragmentary form. 

\end{document}